\pdfoutput=1
\PassOptionsToPackage{table,HTML}{xcolor}

\documentclass[11pt]{article}

\usepackage{ACL2023}

\usepackage{times}
\usepackage{latexsym}
\usepackage[T1]{fontenc}
\usepackage[utf8]{inputenc}
\usepackage{multirow}
\usepackage{graphicx}
\usepackage{arydshln}
\usepackage{booktabs}
\usepackage{amsmath}
\usepackage{subfigure}
\usepackage{subcaption}
\usepackage{diagbox}
\usepackage{float}
\usepackage{enumitem}
\usepackage{microtype}
\usepackage{inconsolata}
\usepackage{titlesec}
\usepackage{bbm}
\usepackage{fontawesome}

\definecolor{myblue}{HTML}{88DEEA}   
\definecolor{myorange}{HTML}{FCA955} 

\newcommand{\C}[1]{\cellcolor{myblue!\fpeval{min(100, #1 * 200)}}+#1}

\newcommand{\CB}[1]{\cellcolor{myblue!\fpeval{min(100, #1 * 200)}}\textbf{+#1}}

\newcommand{\R}[1]{\cellcolor{myorange!\fpeval{min(100, #1 * 25)}}#1$\times$}

\title{Popular but Wrong: Understanding and Mitigating LLM Overconfidence through Knowledge Popularity}

\author{Shiyu Ni\textsuperscript{\rm{1,2,3}}\space\space
Keping Bi\textsuperscript{\rm{1,2,3}}\thanks{~~Corresponding author}\space\space
Jiafeng Guo\textsuperscript{\rm{1,2,3}}\space\space
Xueqi Cheng\textsuperscript{\rm{1,2,3}}\\
    \textsuperscript{\rm 1}State Key Laboratory of AI Safety\\
    \textsuperscript{\rm 2}Institute of Computing Technology, Chinese Academy of Sciences\\
    \textsuperscript{\rm 3}University of Chinese Academy of Sciences\\
    \{nishiyu23z, bikeping, guojiafeng, cxq\}@ict.ac.cn\\
    \href{https://github.com/ShiyuNee/Knowledge-Popularity-for-LLM-Knowledge-Boundary-Perception}{\faGithub~Code} 
}
\begin{document}
\maketitle

\begin{abstract}
Large language models (LLMs) often produce incorrect answers with high confidence, yet the factors associated with such overconfidence remain insufficiently understood. We study this problem through the lens of knowledge popularity. Using entity-centric factual QA derived from Wikidata triplets, we characterize popularity through question entity popularity, answer popularity, and question–answer co-occurrence. We find two consistent patterns. First, hallucinated answers are far from random: compared with ground-truth answers, they tend to be more popular or more frequently associated with the question entity. Second, confidence is strongly tied to the popularity of generated answers: even among incorrect predictions, more popular answers or those with higher question–answer co-occurrence receive higher confidence. Together, these findings suggest that popular but wrong alternatives may contribute to overconfidence. We further show that popularity-related signals can mitigate overconfidence and improve overall confidence estimation. Across six models and three datasets, incorporating knowledge popularity reduces average confidence on incorrect answers from 0.765 to 0.254 and overall ECE from 0.356 to 0.050, while improving Alignment from 77.08\% to 83.72\%. \looseness=-1

\end{abstract}

\section{Introduction \label{sec:intro}}

A trustworthy model should be well calibrated: its confidence in an answer, measured by the generation probability assigned to that answer, should reflect the likelihood that the answer is correct.
However, large language models (LLMs) have been shown to be overconfident, often assigning high confidence to incorrect answers~\citep{jiang2021can,xiong2023can,ni2024llms,yang2023alignment}.
Such miscalibrated confidence poses significant risks, especially in safety-critical domains such as healthcare and finance.
Despite extensive evidence of LLM overconfidence, why incorrect answers can receive high confidence remains insufficiently understood.

\begin{figure}
    \centering
    \includegraphics[width=1.0\linewidth]{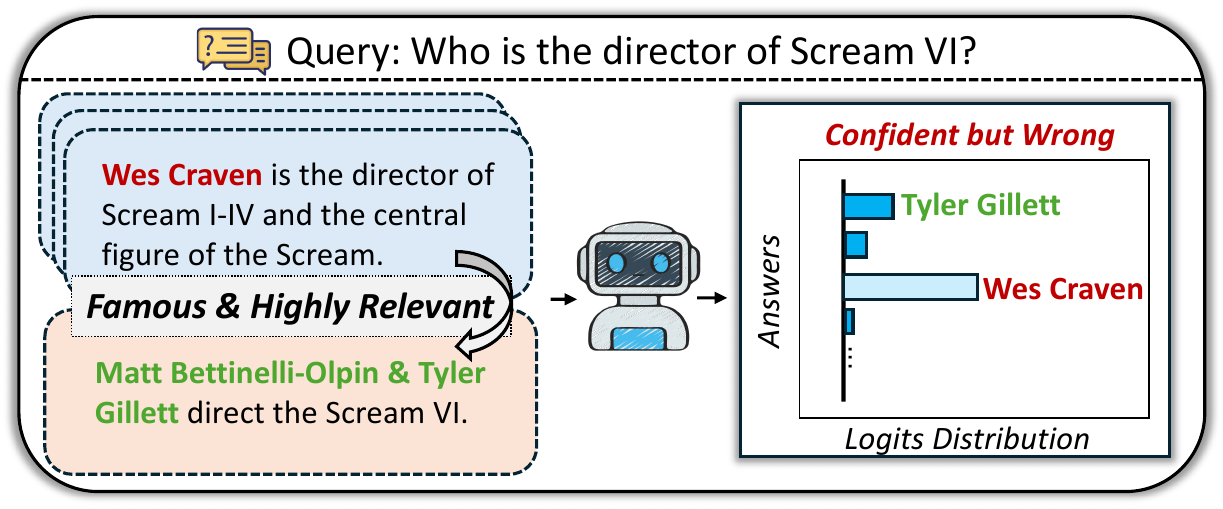}
    \caption{An example where the model generates an incorrect answer that is more popular than the correct answer, while assigning it high confidence.}
    \label{fig:overview}
\end{figure}

Overconfidence can be viewed as a mismatch between the correctness of a generated answer and the confidence assigned to it. We suspect that one source of this mismatch lies in the frequency structure of the knowledge learned by LLMs. Repeated exposure to an entity may make it more likely to be generated, but such popularity does not guarantee that the entity is correct for the queried relation. This motivates us to study overconfidence through the lens of knowledge popularity.\looseness=-1

Knowledge popularity has been widely studied as an important factor in LLM factual recall.
Prior work shows that LLMs are more likely to answer correctly when the required knowledge appears more frequently in the corpus~\citep{mallen-etal-2023-trust,kandpal2023large}.
However, these studies mainly focus on the ground-truth side, explaining when models are likely to know the correct answer.
Overconfidence, in contrast, concerns the confidence assigned to the generated answer, especially when that answer is incorrect.
We therefore shift attention from ground-truth-side popularity to generated-answer-side popularity signals.
Specifically, we ask two questions:
(1) When LLMs make mistakes, what kinds of answers do they generate?
(2) How are generated-answer-side popularity signals associated with model confidence, especially among incorrect answers? \looseness=-1

To answer these questions, we focus on factoid question answering (QA), which allows us to characterize generation-side popularity through question entity popularity, generated answer popularity, and question-answer co-occurrence.
We employ three representative datasets constructed by prior work~\citep{yuksekgonul2023attention}, with examples shown in Figure~\ref{fig:questions}.
As the models' training data are not accessible, we follow prior studies~\citep{mallen-etal-2023-trust,kandpal2023large,yuksekgonul2023attention} and use external statistics as proxies, including Wikidata sitelinks for entity popularity and Wikipedia document co-occurrence for question-answer co-occurrence frequency.\looseness=-1

Our results reveal a systematic popularity-related pattern behind overconfident errors.
When factual recall fails, LLMs do not generate arbitrary wrong answers; instead, their errors tend to gravitate toward entities that are more popular or co-occur more frequently with the question entity than the ground-truth answer.
More importantly, popularity is also related to how strongly models trust these errors: even among incorrect predictions, answers with stronger popularity signals tend to receive higher confidence.
\textit{Together, these findings suggest that popular but wrong alternatives may contribute to overconfidence.}
The pattern is particularly evident when models struggle with the task and persists across model scales. \looseness=-1

Although popular incorrect answers can receive high confidence, correct predictions typically exhibit even stronger and more reliable question-answer association patterns.
We therefore leverage these signals for confidence calibration.
Across six models and three datasets, incorporating knowledge popularity substantially mitigates overconfidence, reducing the average confidence on incorrect answers from 0.765 to 0.254, while improving overall confidence calibration, with ECE decreasing from 0.356 to 0.050 and Alignment increasing from 77.08\% to 83.72\%.
To reduce reliance on external corpora, we further explore using the model itself to estimate popularity signals, which remains a reasonable alternative.\looseness=-1

\begin{figure*}
    \centering
    \includegraphics[width=1.0\linewidth]{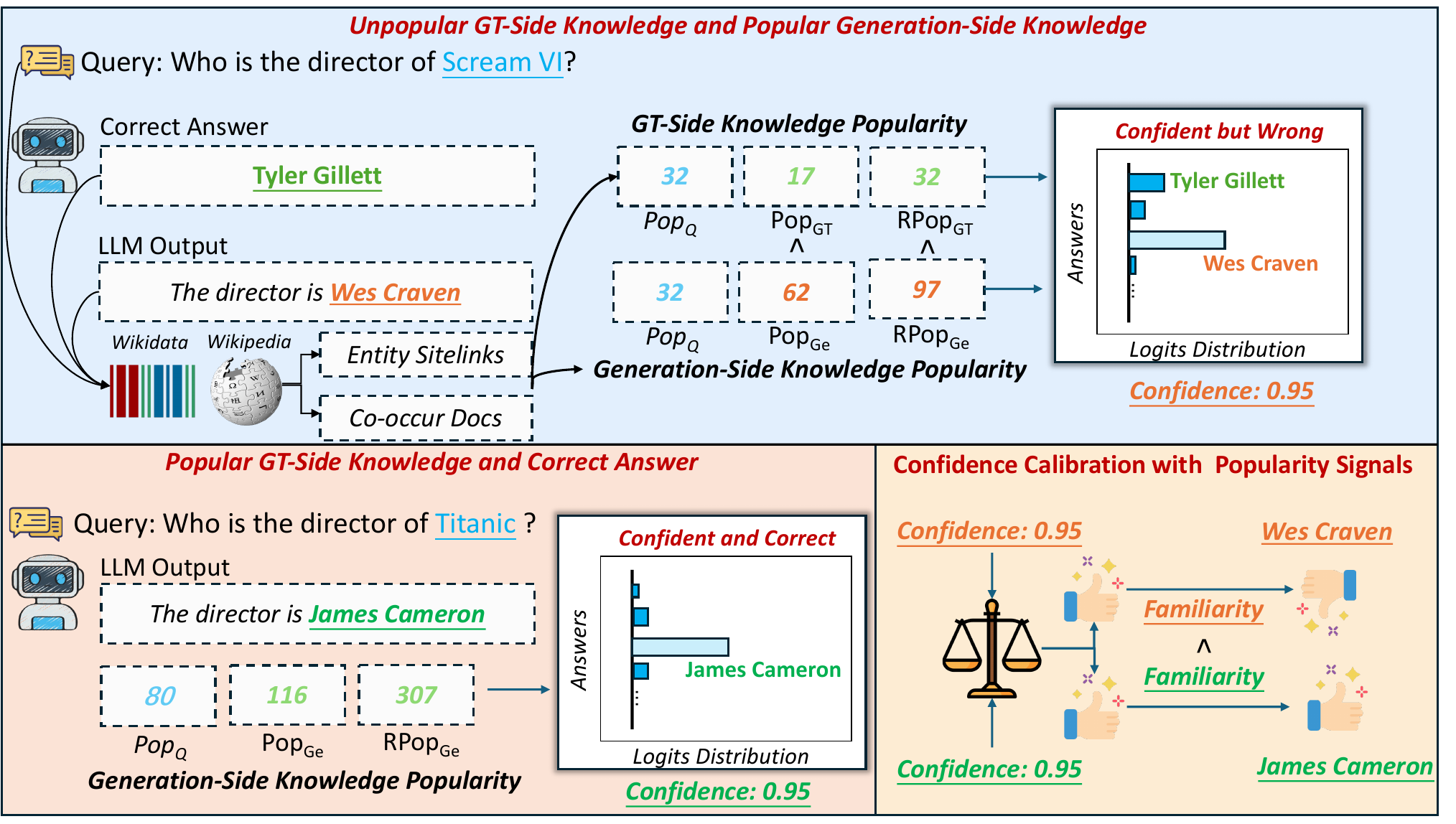}
    \caption{
    Overview of knowledge popularity, overconfidence, and calibration.
    The upper panel illustrates overconfident errors involving popular or frequently co-occurring incorrect answers; the lower panels show correct high-confidence predictions and popularity-aware confidence calibration.
    }
    \label{fig:paper_overview}
\end{figure*}

\section{Related Work \label{sec:related work}}
Existing research on model knowledge boundary perception focuses on assessing model confidence and can be mainly classified into four categories.

\paragraph{Probabilistic Confidence.} 
This line of research uses the probability assigned to an answer as a measure of model confidence~\citep{guo2017calibration, desai2020calibration, jiang2021can, kadavath2022language, si2022prompting, kuhn2023semantic}.
\citet{guo2017calibration} examined modern neural networks (e.g., ResNet~\citep{he2016deep}) and found them to be overconfident, proposing temperature scaling as a remedy.
Later, \citet{desai2020calibration} showed that BERT-style models tend to be relatively well-calibrated, while \citet{jiang2021can} found that pre-trained language models such as T5~\citep{raffel2020exploring} remained overconfident.
More recent work has turned to LLMs, with studies showing that they, too, exhibit overconfidence~\citep{si2022prompting, lin2022teaching, tian2023just}.

\paragraph{Verbalized Confidence.}
LLMs have been shown to express their confidence verbally~\citep{lin2022teaching, yin2023large, tian2023just, xiong2023can, yang2023alignment, ni2024llms}.
Some studies~\citep{yin2023large, ni2024llms} found that LLMs often fail to recognize their knowledge limitations verbally and tend to be overconfident.
\citet{xiong2023can} systematically studied black-box approaches for estimating LLM confidence.
Beyond prompting-based methods, some studies aim to train LLMs to verbalize more accurate confidence~\citep{lin2022teaching, yang2023alignment, zhang2024r}. \looseness=-1

\paragraph{Consistency-based Confidence.}
If the model is confident in its answer, it should maintain consistency across multiple generations. Recent studies have used self-consistency across generations as a proxy for LLM confidence~\citep{manakul2023selfcheckgpt, kuhn2023semantic}.~\citet{zhang2023sac3} extended this by evaluating the consistency of answers across multiple semantically equivalent inputs and across different models. \citet{ding2024retrieve} further adapted this approach to the multilingual setting. 

\paragraph{Confidence Estimation via LLM Internal States.}  
LLMs' internal states have been shown to be effective in evaluating the factuality of their self-generated content~\citep{su2024unsupervised,chen2024inside,wang2024hidden,ni2025towards}. Specifically, \citet{su2024unsupervised} and \citet{chen2024inside} focused on internal states after generation, \citet{wang2024hidden} examined those before generation, and \citet{ni2025towards} explored leveraging
LLMs' internal states to enhance their perception of knowledge boundaries from efficiency and risk perspectives.

We focus on probabilistic confidence for the following reasons:
1) Both the model’s generation probabilities and its knowledge acquisition arise from the same training objective and are expected to align with each other.
2) Models without specialized training often struggle to verbalize confidence accurately~\citep{ni2024large}; consistency-based methods require multiple generations and incur high inference costs; and internal-state-based approaches require access to hidden representations and additional training.
In contrast, probabilistic confidence is readily accessible and has been shown to perform well, especially when answers are short~\citep{ding2024retrieve}.

\vspace{-0.1cm}
\section{Task Formulation \label{sec:task}}
\begin{figure}
    \centering
    \includegraphics[width=1.0\linewidth]{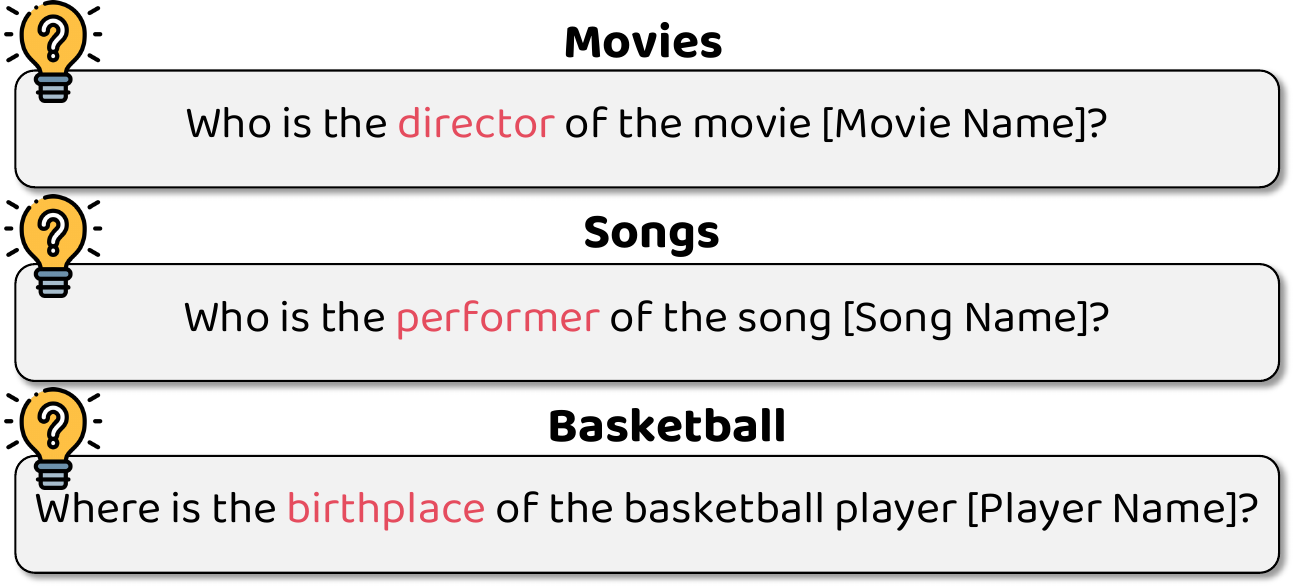}
    \caption{Question examples for each dataset.}
    \label{fig:questions}
\end{figure}

\paragraph{Entity-Centric QA.} We focus on entity-centric knowledge because it allows us to measure knowledge popularity through entities. In entity-centric QA, questions and answers are derived from knowledge triples in the form of (subject, relation, object), where the question queries the relation of a given subject, and the model is expected to generate the corresponding object. Examples of knowledge triples are provided in Table~\ref{tab:Datasets}, with their transformed question forms shown in Figure~\ref{fig:questions}.

\vspace{-0.2cm}
\paragraph{LLM Knowledge Boundary Perception.} The model’s perception of its knowledge boundaries is evaluated by the alignment between its confidence and actual QA performance. QA performance is measured by whether the generated answer contains the ground-truth answer, and confidence is reflected in the generation probabilities of the answer tokens. Specifically, for a question $q$ and a model $M$, the confidence $c$ is computed as $c = \frac{1}{T}\sum_{i=1}^TP(g_i|g_{ < i}),$
where $\{g_1, \dots, g_{T}\}$ are the generated tokens.

\section{Experimental Setup\label{sec:exp-setup}}
\paragraph{Datasets.} \citet{yuksekgonul2023attention} constructed entity-centric QA datasets from Wikidata\footnote{\url{https://query.wikidata.org/sparql}}, using sitelinks as a proxy for entity popularity, which correlates strongly with QA performance. Following their setup, we use three datasets—Movies, Songs, and Basketball—to capture different levels of knowledge popularity. Specifically, question popularity decreases from Movies to Basketball, while ground-truth answer popularity shows the opposite trend. Table~\ref{tab:Datasets} summarizes the knowledge triplets and data statistics, and Figure~\ref{fig:questions} shows examples. \looseness=-1

\paragraph{Entity Popularity.} Following \citet{mallen-etal-2023-trust, yuksekgonul2023attention}, we define the popularity of an entity by the number of sitelinks it has—i.e., the number of Wikipedia pages in different languages that link to it.

\paragraph{Co-occurrence Count.} As Wikipedia is the primary high-quality source for Wikidata, we estimate relation popularity based on Wikipedia content. Specifically, for each entity pair, we measure relation popularity by counting the number of documents in the Wikipedia dump\footnote{\url{https://huggingface.co/datasets/wikimedia/wikipedia}} in which both entities co-occur.

\begin{table}[h]
    \centering
    \caption{Sample counts for each dataset, along with the corresponding subject, relation, and object types.}
    \label{tab:Datasets}
    \scalebox{0.78}{ 
    \begin{tabular}{lrlll}
        \toprule
        Datasets & Count & Subject & Relation & Object  \\
        \midrule
        Movies & 10,964 & Movie & Directed by & Director   \\
        Songs & 2,157 & Song & Performed by & Performer   \\
        Basketball & 13,309 & Player & Birthplace & City  \\
        \bottomrule
    \end{tabular}
    }
\end{table}

\begin{table}[h]
    \centering
    \caption{Definitions of the popularity notation.}
    \label{tab:Notation}
    \scalebox{0.63}{
    \begin{tabular}{ll}
        \toprule
        Notation & Definition \\
        \midrule
        \cellcolor[HTML]{FFC0CB}$\text{Pop}_\text{Q}$ & Popularity of entities in the question \\
        \cellcolor[HTML]{5FFBF1}$\text{Pop}_\text{GT}$ & Popularity of entities in the ground-truth answer \\
        \cellcolor[HTML]{5FFBF1}$\text{RPop}_\text{GT}$ & Co-occurrence count between question and ground-truth entities \\
        \cellcolor[HTML]{C5D7F3}$\text{Pop}_\text{Ge}$ & Popularity of entities in the generated answer \\
        \cellcolor[HTML]{C5D7F3}$\text{RPop}_\text{Ge}$ & Co-occurrence count between question and generated entities \\
        \bottomrule
    \end{tabular}
    }

    \vspace{-10pt}
\end{table}

\paragraph{LLMs.} We conduct experiments on six representative LLMs: five open-source models—Llama 3-8B-Instruct~\citep{dubey2024llama}, Qwen2-7B-Instruct~\citep{yang2024qwen2}, and the Qwen2.5-Instruct series (7B, 14B, and 32B)~\citep{hui2024qwen2}—as well as a black-box model, ChatGPT (i.e., GPT-3.5-Turbo-1106)~\citep{achiam2023gpt}. We do not include the Qwen2.5 series on Basketball because their accuracy is too low (see Table~\ref{tab:acc_and_conf}).

\paragraph{Answer Generation.} For all the models, we use greedy search, selecting the token with the highest probability at each generation step.

\paragraph{Metrics.} For each question $q_i$, we measure answer correctness using accuracy $acc_i$, where the generated answer is considered correct if it contains the ground-truth answer. The model's confidence $c_i$ is defined as the generation probability of the answer, as described in~\S\ref{sec:task}. 

\paragraph{Partial Correlation.} For each signal, we compute the partial Spearman correlation coefficient to ensure reliable analysis by controlling for the effects of the other signals. For example, when analyzing how popularity influences confidence, we isolate the effect of $\text{Pop}_\text{Ge}$ by controlling for $\text{RPop}_\text{Ge}$ and $\text{Pop}_Q$. Details can be found in~\S~\ref{sec:partial-corr-details}.

\section{What Do LLMs Generate When They Hallucinate?\label{sec:rq1}}

In this section, we first examine the relationship between knowledge popularity and QA performance and then focus on what the models generate when they hallucinate.

\subsection{Ground-Truth-Side Popularity and Factual Recall \label{sec:rq1-learn}}
Unlike previous studies that focus on a single signal~\citep{mallen-etal-2023-trust,kandpal2023large}, we compare the strengths of the correlations between three different signals (i.e., $\text{Pop}_Q$, $\text{Pop}_\text{GT}$, and $\text{RPop}_\text{GT}$) and QA performance. For each signal, we control for the other two and compute its partial correlation with accuracy. The partial Spearman correlation coefficients are shown in Table~\ref{tab:acc-corr}. The findings are:
1) Co-occurrence frequency exhibits the strongest correlation with model QA performance.
2) Relation specificity is also important (see Figure~\ref{fig:relation_quality}).
Due to space limitations, details are provided in~\S~\ref{sec:acc_learning}.

\begin{figure}
    \centering
    \includegraphics[width=1.0\linewidth]{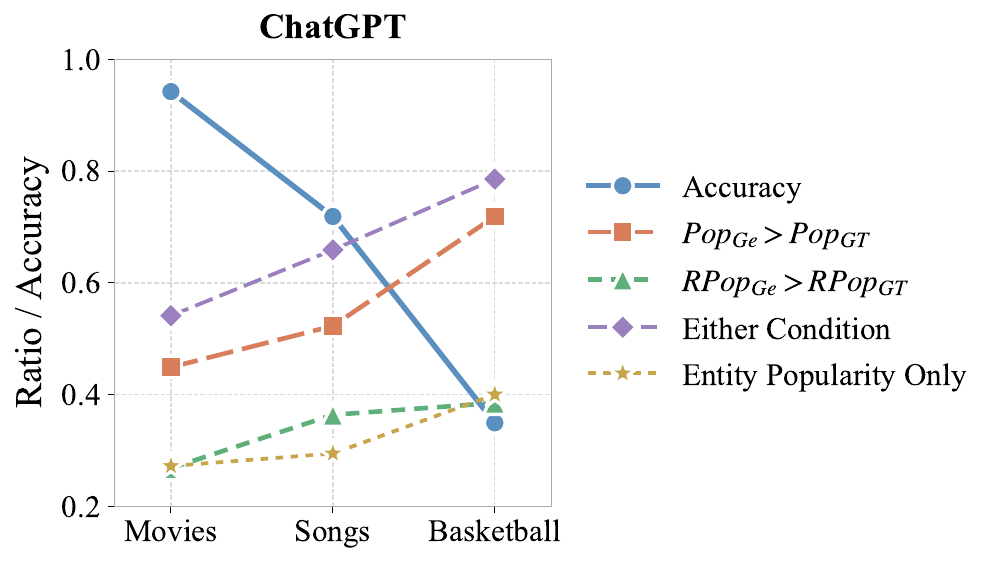}
    \caption{QA accuracy and failure modes across datasets for ChatGPT.
    Accuracy uses all samples, whereas failure-mode ratios use incorrect
    answers only. \textit{Either Condition} means that either popularity
    comparison holds; \textit{Entity Popularity Only} means that only
    $Pop_{Ge}>Pop_{GT}$ holds.}
    \label{fig:failure-mode}
    \vspace{-0.2cm}
\end{figure}

\subsection{Hallucinated Answers Are More Popular or Strongly Associated \label{sec:failuremode}}

We analyze the popularity of the entities generated by the model when it makes mistakes, as well as their co-occurrence frequency with the question entity, and compare them with the ground-truth answers. The results are shown in Figure~\ref{fig:failure-mode} and Table~\ref{tab:chatgpt-failure-mode}. Due to space limitations, the full results are provided in Table~\ref{tab:cross-dataset-all-models}. Our conclusions are as follows.


\begin{table}[t]
\caption{Median popularity of ground-truth and generated answers on ChatGPT, grouped by correct and incorrect predictions. CV denotes the coefficient of variation, and Answer Unique Ratio is the fraction of unique generated answers among incorrect samples.}
\centering
\small
\setlength{\tabcolsep}{6pt}
\begin{tabular}{l c c c}
\toprule
\textbf{Metric} & \textbf{Movies} & \textbf{Songs} & \textbf{Basketball} \\
\midrule
\multicolumn{4}{l}{\textit{Correct Answers}} \\
\midrule
$\text{Pop}_{\text{Q}}$ (median) & 22.0 & 10.0 & 9.0 \\
$\text{Pop}_{\text{GT}}$ (median) & 28.0 & 64.0 & 132.0 \\
$\text{RPop}_{\text{GT}}$ (median) & 9.0 & 17.0 & 3.0 \\
\midrule
\multicolumn{4}{l}{\textit{Incorrect Answers -- Ground Truth}} \\
\midrule
$\text{Pop}_{\text{Q}}$ (median) & 18.0 & 7.0 & 7.0 \\
$\text{Pop}_{\text{GT}}$ (median) & 21.0 & 31.5 & 75.0 \\
$\text{RPop}_{\text{GT}}$ (median) & 5.0 & 7.0 & 1.0 \\
\midrule
\multicolumn{4}{l}{\textit{Incorrect Answers -- Generated}} \\
\midrule
$\text{Pop}_{\text{Ge}}$ (median) & 5.0 & 39.0 & 186.0 \\
$\text{RPop}_{\text{Ge}}$ (median) & 2.0 & 4.5 & 1.0 \\
$\text{Pop}_{\text{Ge}}$ CV & 1.58 & 0.90 & 0.58 \\
Answer Unique Ratio & 0.787 & 0.821 & 0.133   \\

\bottomrule
\end{tabular}

\label{tab:chatgpt-failure-mode}
\end{table}

\paragraph{LLMs tend to generate more familiar entities when they hallucinate.}
As shown in Figure~\ref{fig:failure-mode}, among the incorrectly answered samples, approximately 40\% to 70\% of the generated entities are more popular than the ground-truth entities, while 20\% to 40\% of the generated entities co-occur with the question entities more frequently than the correct answers do. Together, these two categories account for the majority of the incorrect samples.
The former suggests that the model may increase the generation probability of an entity simply because it is more popular, assigning it a higher probability than the correct answer. For example, when asked about the director of a movie, the model may not know the correct answer and instead output a better-known director.
The latter suggests that the model can also be misled by entities that frequently co-occur with the question entity. For instance, it may answer with the producer of a movie instead of the director. \looseness=-1

\paragraph{Substantial domain knowledge gaps are associated with stronger popularity-biased generation.}
As shown in Figure~\ref{fig:failure-mode}, both popularity-bias ratios are
substantially higher on Basketball, where model accuracy is lowest. Although
the differences between Movies and Songs are modest, the contrast with
Basketball suggests that, when models have substantially less knowledge
about a domain, their hallucinations are more likely to favor answers that
are globally popular or strongly associated with the question. 
Table~\ref{tab:chatgpt-failure-mode} reports detailed popularity and
question-answer association statistics for the ground-truth and generated
answers, together with the CV and answer unique ratio to characterize the
concentration of generated-answer distributions.

\looseness=-1

\begin{table*}[h!]
\centering
\caption{Partial Spearman correlations between question--generated-answer popularity and confidence. Bold marks the strongest positive predictor of confidence on incorrect answers. The ratio denotes how many times larger the correlation on incorrect samples is than on correct samples. Ratios involving sign changes or near-zero denominators are marked ``---'' as they are not meaningful.}
\small
\setlength{\tabcolsep}{3pt}
\scalebox{0.85}{\begin{tabular}{l l cccc cccc cccc}
\toprule
& & \multicolumn{4}{c}{$\text{RPop}_\text{Ge}$} & \multicolumn{4}{c}{$\text{Pop}_\text{Q}$} & \multicolumn{4}{c}{$\text{Pop}_\text{Ge}$} \\
\cmidrule(lr){3-6}\cmidrule(lr){7-10}\cmidrule(lr){11-14}
Datasets & Models & Total & Correct & Incorrect & Ratio & Total & Correct & Incorrect & Ratio & Total & Correct & Incorrect & Ratio \\
\midrule
\multirow{6}{*}{Movies}
 & Llama-3-8B   & \C{0.541} & \C{0.262} & \C{0.267} & \R{1.0} & \C{0.135} & \C{0.164} & \C{0.045} & \R{0.3} & $-$0.014 & \C{0.075} & \CB{0.186} & \R{2.5} \\
 & Qwen2-7B     & \C{0.541} & \C{0.188} & \C{0.179} & \R{1.0} & \C{0.104} & \C{0.167} & \C{0.046} & \R{0.3} & \C{0.112} & \C{0.158} & \CB{0.263} & \R{1.7} \\
 & ChatGPT      & \C{0.172} & \C{0.109} & \CB{0.499} & \R{4.6} & \C{0.092} & \C{0.082} & \C{0.060} & \R{0.7} & \C{0.045} & \C{0.078} & $-$0.112 & --- \\
 & Qwen2.5-7B   & \C{0.419} & \C{0.178} & \C{0.195} & \R{1.1} & \C{0.107} & \C{0.172} & \C{0.051} & \R{0.3} & \C{0.239} & \C{0.156} & \CB{0.310} & \R{2.0} \\
 & Qwen2.5-14B  & \C{0.580} & \C{0.287} & \C{0.219} & \R{0.8} & \C{0.098} & \C{0.142} & \C{0.053} & \R{0.4} & \C{0.103} & \C{0.053} & \CB{0.262} & \R{4.9} \\
 & Qwen2.5-32B  & \C{0.596} & \C{0.254} & \C{0.229} & \R{0.9} & \C{0.104} & \C{0.133} & \C{0.046} & \R{0.3} & \C{0.091} & \C{0.074} & \CB{0.242} & \R{3.3} \\
\midrule
\multirow{6}{*}{Songs}
 & Llama-3-8B   & \C{0.609} & \C{0.163} & \CB{0.458} & \R{2.8} & \C{0.107} & \C{0.158} & \C{0.105} & \R{0.7} & \C{0.026} & $-$0.004 & \C{0.012} & --- \\
 & Qwen2-7B     & \C{0.375} & \C{0.065} & \CB{0.140} & \R{2.2} & \C{0.146} & \C{0.076} & \CB{0.084} & \R{1.1} & \C{0.071} & \C{0.048} & \CB{0.105} & \R{2.2} \\
 & ChatGPT      & \C{0.375} & \C{0.181} & \CB{0.656} & \R{3.6} & \C{0.037} & \C{0.050} & \C{0.011} & \R{0.2} & \C{0.236} & \C{0.255} & $-$0.014 & --- \\
 & Qwen2.5-7B   & \C{0.357} & \C{0.011} & \CB{0.221} & --- & \C{0.059} & \C{0.002} & \C{0.003} & --- & \C{0.158} & \C{0.309} & \C{0.205} & \R{0.7} \\
 & Qwen2.5-14B  & \C{0.451} & \C{0.080} & \CB{0.253} & \R{3.1} & \C{0.063} & \C{0.067} & $-$0.008 & --- & \C{0.161} & \C{0.243} & \C{0.168} & \R{0.7} \\
 & Qwen2.5-32B  & \C{0.441} & \C{0.059} & \CB{0.237} & \R{4.0} & \C{0.131} & \C{0.126} & \C{0.067} & \R{0.5} & \C{0.149} & \C{0.147} & \CB{0.194} & \R{1.3} \\
\midrule
\multirow{3}{*}{Basketball}
 & Llama-3-8B   & $-$0.039 & $-$0.046 & $-$0.102 & \R{2.2} & \C{0.175} & \C{0.214} & \CB{0.158} & \R{0.7} & $-$0.007 & $-$0.177 & $-$0.016 & --- \\
 & Qwen2-7B     & \C{0.008} & \C{0.026} & $-$0.024 & --- & \C{0.146} & \C{0.015} & \CB{0.168} & \R{11.4} & \C{0.125} & $-$0.009 & \CB{0.099} & --- \\
 & ChatGPT      & \C{0.188} & \C{0.179} & \C{0.040} & \R{0.2} & \C{0.255} & \C{0.242} & \CB{0.161} & \R{0.7} & $-$0.179 & $-$0.136 & $-$0.100 & \R{0.7} \\
\bottomrule
\end{tabular}}
\label{tab:conf-split}
\end{table*}

\section{What Drives Confidence in Generated Answers? \label{sec:rq2}}
In this section, we investigate the relationship between model confidence and the popularity signals associated with the question and generated answer ($\text{Pop}_Q$, $\text{Pop}_\text{Ge}$, and $\text{RPop}_\text{Ge}$). The results are shown in Table~\ref{tab:conf-split}. Overall (Total column), we find that \textbf{confidence is driven by question--generated-answer popularity}, especially $\text{RPop}_\text{Ge}$. We further analyze correctly and incorrectly answered samples separately and summarize the findings as follows.

\begin{figure}
    \centering
    \includegraphics[width=1.0\linewidth]{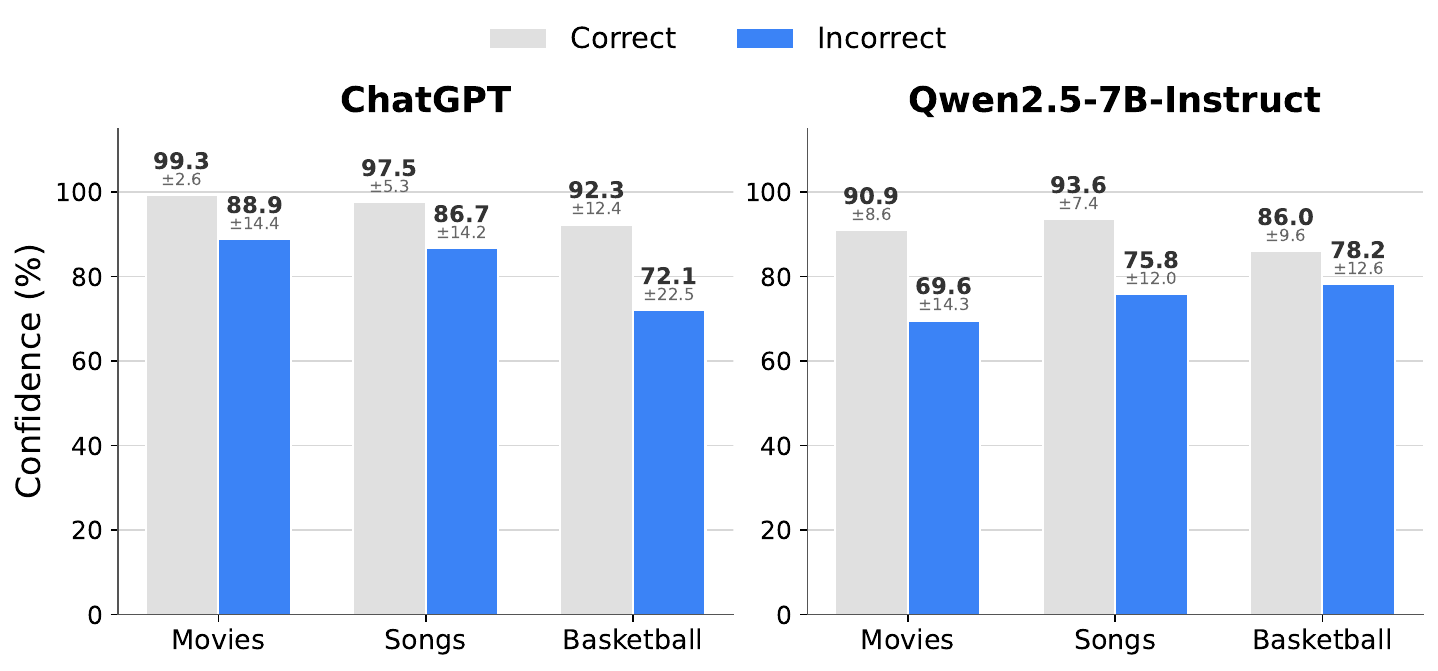}
    \caption{Mean confidence and its variation for correctly and incorrectly answered samples.}
    \label{fig:confidence-wrong-right}
\end{figure}

\paragraph{Within correctly answered samples, uniformly high popularity and confidence lead to weak observable correlations.}
Although Table~\ref{tab:conf-split} shows weak correlations between popularity signals and confidence for correctly answered samples, this does not imply independence. Correct answers typically involve highly popular and strongly associated knowledge (see Table~\ref{tab:chatgpt-failure-mode}), driving confidence to consistently high levels with limited variance (Figure~\ref{fig:confidence-wrong-right}). When both popularity and confidence are uniformly high, their correlation is naturally hard to detect. This also helps explain why LLMs rarely exhibit conservativeness~\citep{ni2024llms,xiong2023can}: correctly answered knowledge is generally popular, and models assign high confidence. \looseness=-1

\paragraph{For incorrectly answered samples, model confidence is driven by question–answer pair popularity.}
As shown in Table~\ref{tab:conf-split}, confidence for incorrect answers generally increases with question–answer pair popularity, though the dominant signal varies across datasets. In Movies and Songs, confidence is strongly correlated with $\text{RPop}_\text{Ge}$, while this effect disappears in Basketball. Moreover, the correlation between confidence and $\text{Pop}_\text{Ge}$ weakens from Movies to Songs and is weakest in Basketball.
As discussed in~\S~\ref{sec:failuremode}, LLMs increasingly generate more popular and concentrated entities from Movies to Songs to Basketball. As entity popularity becomes uniformly high, selecting a more popular entity provides diminishing additional gain in confidence, making confidence less sensitive to $\text{Pop}_\text{Ge}$.
Basketball shows a different pattern: due to extremely sparse player–city co-occurrence (see Table~\ref{tab:chatgpt-failure-mode}), even higher $\text{RPop}_\text{Ge}$ values remain small in absolute terms, so popularity differences lead to only marginal changes in confidence. \looseness=-1

\paragraph{Scaling cannot mitigate the model’s higher confidence in popular but incorrect answers.}
As shown in Table~\ref{tab:conf-split}, within the Qwen2.5 model series, increasing model scale does not weaken the correlation between confidence and $\text{RPop}_\text{Ge}$ for incorrectly answered samples in the Movies and Songs datasets. Similarly, in Basketball, confidence and $\text{Pop}_Q$ remain high even for incorrect samples.

\textit{Taken together, these findings suggest a popularity-related pathway to overconfidence: when the correct question-answer association is weak, LLMs tend to generate more familiar but incorrect alternatives, and these alternatives also tend to receive high confidence.}

\section{Calibrating Confidence with Popularity Signals \label{sec:rq3}}

Although LLMs tend to generate more common answers when making errors, $\text{RPop}_\text{Ge}$ remains substantially lower than in correctly answered samples (see Table~\ref{tab:delta-gcoo}). This motivates us to mitigate overconfidence using knowledge-popularity signals. \looseness=-1

\subsection{Popularity-Aware Confidence Estimation}

For a given question $q$, let the model's response be $r$ and its confidence be $c$. We leverage a popularity signal $p$ to map $(p, c)$ to $y$, where $y$ denotes the ground-truth confidence (i.e., the correctness of the answer). We consider three types of popularity signals: $\text{Pop}_\text{Q}$, $\text{Pop}_\text{Ge}$, and $\text{RPop}_\text{Ge}$, as well as a combined setting that uses all three signals.
To reduce reliance on external corpora, we also explore using the LLM itself to estimate each popularity signal, denoted as $\widehat{\mathrm{Pop}}_{\mathrm{Q}}$, $\widehat{\mathrm{Pop}}_{\mathrm{Ge}}$, and $\widehat{\mathrm{RPop}}_{\mathrm{Ge}}$. \looseness=-1

We learn a popularity-aware confidence estimator using a lightweight MLP, as defined below:
\begin{equation}
    P(\hat{y}=1) = \sigma\left( \text{MLP} \left( x  \right) \right),
\end{equation}
where $\sigma$ denotes the sigmoid function, $x \in \mathbbm{R}^{d}$ represents the input features, and $d$ is the number of input features (e.g., $d=2$ for $c+\text{Pop}_\text{Q}$). The MLP has three hidden layers with 128, 64, and 32 units, each followed by batch normalization, ReLU activation, and dropout (rate 0.3).
We employ cross-entropy loss as the training objective:
\begin{equation}
\mathcal{L}_{\mathrm{CE}}
= -\sum_{i=1}^{N}
\left[y_i \log P_i + (1-y_i)\log(1-P_i)\right],
\end{equation}
where $y_i$ is the ground-truth correctness for the $i$-th training sample, $N$ is the number of training samples, and $P_i$ denotes $P(\hat{y}_i = 1)$. Detailed training parameters can be found in~\S~\ref{sec:training-details}.

\begin{table*}[t]
\centering
\small
\setlength{\tabcolsep}{3pt}
\caption{Conf$_\text{w}$ (average confidence on incorrect samples), ECE, and Alignment averaged over six models. For Align., bold/underline = highest/lowest; for Conf$_\text{w}$ and ECE, bold/underline = lowest/second-lowest. ``All'' denotes the use of all three types of popularity signals.}
\label{tab:detect}
\begin{tabular}{l ccc ccc ccc ccc}
\toprule
& \multicolumn{3}{c}{Movies} & \multicolumn{3}{c}{Songs} & \multicolumn{3}{c}{Basketball} & \multicolumn{3}{c}{Avg.} \\
\cmidrule(lr){2-4}\cmidrule(lr){5-7}\cmidrule(lr){8-10}\cmidrule(lr){11-13}
Method & Conf$_\text{w}$ & ECE & Align. & Conf$_\text{w}$ & ECE & Align. & Conf$_\text{w}$ & ECE & Align. & Conf$_\text{w}$ & ECE & Align. \\
\midrule
\multicolumn{13}{l}{\textit{Confidence and calibration baselines}} \\
\quad Prob             & 0.780 & 0.387 & 81.88 & 0.775 & 0.355 & 80.72 & 0.741 & 0.326 & 68.66 & 0.765 & 0.356 & 77.08 \\
\quad Consis           & 0.320 & 0.157 & 79.33 & 0.295 & 0.121 & 83.52 & \textbf{0.336} & 0.173 & 67.16 & 0.317 & 0.150 & 76.67 \\
\quad Verbal           & 0.497 & 0.350 & \underline{64.97} & 0.584 & 0.341 & 65.85 & 0.388 & 0.420 & 57.95 & 0.490 & 0.371 & \underline{62.92} \\
\quad Prob + Temp      & 0.603 & 0.247 & 81.88 & 0.595 & 0.218 & 80.72 & 0.571 & 0.147 & 68.66 & 0.590 & 0.204 & 77.08 \\
\quad Prob + Platt     & 0.307 & 0.082 & 81.88 & 0.398 & 0.173 & 80.72 & 0.414 & 0.060 & 68.66 & 0.373 & 0.105 & 77.08 \\
\quad Prob + Isotonic  & 0.258 & 0.024 & 81.88 & 0.269 & \textbf{0.055} & 80.72 & 0.388 & \underline{0.035} & 68.66 & 0.305 & \textbf{0.038} & 77.08 \\
\midrule
\multicolumn{13}{l}{\textit{Corpora-based popularity}} \\
\quad $\text{Pop}_\text{Q}$  & $-$ & $-$ & 70.27 & $-$ & $-$ & 68.31 & $-$ & $-$ & \underline{55.02} & $-$ & $-$ & 64.53 \\
\quad $\text{Pop}_\text{Ge}$ & $-$ & $-$ & 67.66 & $-$ & $-$ & \underline{63.20} & $-$ & $-$ & 61.42 & $-$ & $-$ & 64.09 \\
\quad $\text{RPop}_\text{Ge}$ & $-$ & $-$ & 87.64 & $-$ & $-$ & 87.26 & $-$ & $-$ & 62.04 & $-$ & $-$ & 78.98 \\
\quad Prob + $\text{Pop}_\text{Q}$  & 0.257 & 0.024 & 83.17 & 0.315 & 0.102 & 81.75 & 0.395 & 0.037 & 68.71 & 0.322 & 0.054 & 77.88 \\
\quad Prob + $\text{Pop}_\text{Ge}$ & 0.260 & 0.025 & 82.73 & 0.340 & 0.078 & 81.92 & \underline{0.382} & 0.049 & 71.57 & 0.327 & 0.050 & 78.74 \\
\quad Prob + $\text{RPop}_\text{Ge}$ & \underline{0.180} & 0.024 & 89.54 & \textbf{0.219} & 0.084 & \textbf{88.42} & 0.395 & 0.045 & 69.87 & \underline{0.265} & 0.051 & 82.61 \\
\quad Prob + All Pop.\ & \textbf{0.145} & \textbf{0.021} & \textbf{91.40} & \underline{0.236} & 0.095 & 87.97 & 0.382 & \textbf{0.035} & \textbf{71.79} & \textbf{0.254} & 0.050 & \textbf{83.72} \\
\midrule
\multicolumn{13}{l}{\textit{Self-generated popularity}} \\
\quad Prob + $\widehat{\text{RPop}}_\text{Ge}$ & 0.259 & \underline{0.023} & 83.02 & 0.318 & 0.078 & 82.77 & 0.405 & 0.042 & 68.66 & 0.327 & \underline{0.048} & 78.15 \\
\quad Prob + All Pop.   & 0.246 & 0.024 & 83.49 & 0.312 & \underline{0.075} & 83.57 & 0.384 & 0.048 & 69.99 & 0.314 & 0.049 & 79.02 \\
\bottomrule
\end{tabular}
\end{table*}

\subsection{Experimental Setup}
\label{sec:rq3-detect}

\paragraph{Evaluation Metrics.} 
To evaluate overconfidence mitigation and calibration quality, we report two calibration metrics: 1) \textbf{Conf$_\text{w}$}, the average confidence on incorrect samples, which directly measures overconfidence; 2) \textbf{Expected Calibration Error (ECE)}~\citep{guo2017calibration}, which measures the discrepancy between predicted confidence and empirical accuracy. Beyond calibration quality, we additionally evaluate whether calibrated confidence better supports confidence-based binary decisions. 
We report \textbf{Alignment}, defined as the fraction of test samples whose binarized confidence prediction matches the ground-truth correctness. See Appendix~\ref{sec:training-details} for full details.

\paragraph{Baselines.} We consider the following representative baseline methods. We exclude approaches that require access to internal model states, as they are not applicable when only output token probabilities are available.

\begin{itemize}[leftmargin=*,topsep=2pt,itemsep=1pt]
\item \textbf{Token probability} ({Prob})~\citep{jiang2021can}: the mean probability of generated tokens.

\item \textbf{Self-consistency} ({Consis})~\citep{manakul2023selfcheckgpt}: estimates confidence via semantic consistency among multiple sampled responses. If the model knows the correct answer, its outputs should be consistent. We sample 10 additional responses (temperature 1.0); for ChatGPT, we use 3 due to cost constraints. Consistency is computed using Qwen2.5-32B-Instruct.

\item \textbf{Verbalized confidence} ({Verbal})~\citep{yin2023large}: prompts the model to explicitly judge whether it can answer the question correctly.

\item \textbf{Temperature scaling} ({Temp})~\citep{guo2017calibration}: learns a single scalar temperature to rescale token-level probabilities for calibration.

\item \textbf{Platt scaling} ({Platt})~\citep{platt1999probabilistic}: fits a logistic mapping from token probability to answer correctness.

\item \textbf{Isotonic regression} ({Isotonic})~\citep{zadrozny2002transforming}: learns a non-parametric monotonic mapping from token probability to answer correctness.
\end{itemize}

\paragraph{Training and evaluation protocol.}
All methods are evaluated independently for each (dataset, model) pair. To prevent extreme QA accuracy on a dataset from causing Alignment to be unduly influenced by the majority class, we balance each (dataset, model) pair whose accuracy falls outside $[0.25, 0.75]$ by undersampling the majority class to a 1:1 ratio. We then apply a 50/50 stratified train/test split (seed 42). All parameter fitting and threshold selection use only the training split, and the held-out test split is used solely for evaluation. For MLP-based methods, we select the checkpoint using a 10\% validation subset drawn from the training split. Full details are provided in Appendix~\ref{sec:training-details}.

\subsection{Results and Analysis}

Table~\ref{tab:detect} reports calibration and alignment results averaged over six models.

\paragraph{Popularity signals substantially mitigate overconfidence.}
Adding popularity signals greatly reduces confidence on incorrect samples. 
Compared with vanilla Prob (Conf$_\text{w}$=0.765), Prob + $\mathrm{RPop}_{\mathrm{Ge}}$ lowers Conf$_\text{w}$ to 0.265 on average, while combining all popularity signals further reduces it to 0.254. 
The improvement is especially large on Movies and Songs, indicating that popularity-aware calibration effectively suppresses overconfident errors.

\paragraph{Popularity improves both calibration and confidence-based decisions.}
Both standard supervised calibration and popularity-aware methods substantially reduce ECE. Platt scaling and isotonic regression reduce ECE from 0.356 with vanilla Prob to 0.105 and 0.038, respectively, but leave Alignment unchanged at 77.08\%. Popularity-aware methods similarly achieve ECE around 0.05, while Prob + All Pop. improves Alignment from 77.08\% to 83.72\%. This contrast suggests that supervised calibration primarily corrects the probability scale, whereas popularity signals additionally improve discrimination between correct and incorrect answers. We disentangle these contributions in detail in~\S\ref{sec:rq3-further}. \looseness=-1

\paragraph{$\mathrm{RPop}_{\mathrm{Ge}}$ is the most effective standalone signal.}
Among individual popularity features, $\mathrm{RPop}_{\mathrm{Ge}}$ consistently yields the strongest performance. 
As a standalone detector, it already achieves high Alignment on Movies (87.64\%) and Songs (87.26\%). 

\paragraph{Self-generated popularity signals remain effective.}
LLM-generated popularity estimates also improve calibration over confidence-only baselines. 
For example, Prob + All Pop.\ using self-generated signals improves average Alignment from 77.08\% to 79.02\% and reduces ECE from 0.356 to 0.049. 
However, the gains remain smaller than those achieved using external popularity signals.

\subsection{Further Analysis \label{sec:rq3-further}}

We further conduct fine-grained analyses to better understand the source of the calibration gains and their robustness across evaluation settings.
In addition to ECE and Alignment, we report \textbf{Brier score}~\citep{glenn1950verification}, which measures sample-level probabilistic prediction error, and \textbf{AUROC}~\citep{fawcett2006introduction}, which measures discrimination between correct and incorrect answers.\looseness=-1

To isolate the contribution of popularity features from that of supervised nonlinear fitting, we introduce MLP($c$), which takes confidence alone as input while using the same architecture, objective, data split, and optimization procedure as our popularity-aware estimator, MLP($c+\mathrm{RPop}_{\mathrm{Ge}}$).

\begin{table}[t]
\centering
\small
\setlength{\tabcolsep}{3.5pt}
\caption{Fine-grained comparison of calibration baselines and the capacity-matched ablation. The main setting follows the same evaluation protocol as Table~\ref{tab:detect}; the natural setting retains the original class distribution.}
\label{tab:matched-calib}
\begin{tabular}{lcccc}
\toprule
Method & ECE$\downarrow$ & Brier$\downarrow$ & AUROC$\uparrow$ & Align.$\uparrow$ \\
\midrule
\multicolumn{5}{l}{\textit{Balanced class distribution}} \\
Prob & 0.356 & 0.322 & 0.839 & 77.08 \\
Prob + Temp & 0.204 & 0.227 & 0.839 & 77.08 \\
Prob + Platt & 0.105 & 0.172 & 0.839 & 77.08 \\
Prob + Isotonic & \textbf{0.038} & 0.156 & 0.837 & 77.08 \\
MLP($c$) & 0.052 & 0.158 & 0.840 & 77.08 \\
MLP($c+\mathrm{RPop}_{\mathrm{Ge}}$)
& 0.050 & \textbf{0.123} & \textbf{0.877} & \textbf{82.62} \\
\midrule
\multicolumn{5}{l}{\textit{Natural class distribution}} \\
Prob & 0.513 & 0.421 & 0.836 & 86.29 \\
Prob + Temp & 0.345 & 0.258 & 0.836 & 86.29 \\
Prob + Platt & 0.054 & 0.108 & 0.836 & 86.29 \\
Prob + Isotonic & \textbf{0.017} & 0.101 & 0.835 & 86.23 \\
MLP($c$) & 0.031 & 0.103 & 0.835 & 86.29 \\
MLP($c+\mathrm{RPop}_{\mathrm{Ge}}$)
& 0.041 & \textbf{0.081} & \textbf{0.876} & \textbf{89.71} \\
\bottomrule
\end{tabular}
\end{table}

\paragraph{Popularity mainly improves correctness discrimination.}
As shown in Table~\ref{tab:matched-calib}, the standard calibration baselines from Table~\ref{tab:detect}, together with the capacity-matched ablation, show that supervised recalibration already accounts for most of the ECE reduction: MLP($c$) reduces ECE from 0.356 to 0.052, comparable to 0.050 after adding $\mathrm{RPop}_{\mathrm{Ge}}$.
In contrast, MLP($c$) leaves AUROC nearly unchanged (0.839$\rightarrow$0.840) and Alignment at 77.08\%, whereas adding $\mathrm{RPop}_{\mathrm{Ge}}$ improves AUROC to 0.877 and Alignment to 82.62\%.
This suggests that the main additional benefit of popularity lies in improving correctness discrimination across samples, rather than further rescaling confidence values.

\paragraph{Popularity remains effective under the natural class distribution.}
We further evaluate all methods while retaining the original class distributions in both the training and test sets.
As shown in Table~\ref{tab:matched-calib}, MLP($c+\mathrm{RPop}_{\mathrm{Ge}}$) improves AUROC from 0.835 to 0.876 and Alignment from 86.29\% to 89.71\% over MLP($c$).
It also reduces the Brier score from 0.103 to 0.081 while maintaining a low ECE of 0.041.
These results show that the additional discrimination provided by $\mathrm{RPop}_{\mathrm{Ge}}$ also holds under the natural data distribution.

\begin{table}[t]
\centering
\small
\setlength{\tabcolsep}{4pt}
\caption{Cross-model and cross-dataset transfer under the natural class distribution.}
\label{tab:transfer-calib}
\begin{tabular}{lcccc}
\toprule
Method & ECE$\downarrow$ & Brier$\downarrow$ & AUROC$\uparrow$ & Align.$\uparrow$ \\
\midrule
\multicolumn{5}{l}{\textit{Cross-model transfer}} \\
MLP($c$) & 0.114 & 0.126 & 0.830 & 83.21 \\
MLP($c+\mathrm{RPop}_{\mathrm{Ge}}$)
& \textbf{0.106} & \textbf{0.104} & \textbf{0.861} & \textbf{86.00} \\
\midrule
\multicolumn{5}{l}{\textit{Cross-dataset transfer}} \\
MLP($c$) & \textbf{0.137} & 0.144 & 0.828 & 80.37 \\
MLP($c+\mathrm{RPop}_{\mathrm{Ge}}$)
& 0.147 & \textbf{0.139} & \textbf{0.845} & \textbf{81.14} \\
\bottomrule
\end{tabular}
\end{table}

\paragraph{Popularity-aware calibration transfers across models and datasets.}
We further evaluate whether the learned calibration rule generalizes across models and datasets.
For cross-model transfer, adding $\mathrm{RPop}_{\mathrm{Ge}}$ improves AUROC from 0.830 to 0.861 and Alignment from 83.21\% to 86.00\%.
For cross-dataset transfer, which also corresponds to transfer across relation types, AUROC improves from 0.828 to 0.845 and Alignment from 80.37\% to 81.14\%.
Although the gains become smaller under cross-dataset transfer, $\mathrm{RPop}_{\mathrm{Ge}}$ still provides useful transferable information beyond model confidence.
Details are provided in the Appendix~\ref{sec:full-transfer-results}.

\section{Conclusion}
This paper investigates LLM overconfidence in factual QA through the lens of knowledge popularity. We find that hallucinated answers are far from random: incorrect predictions are often more popular than the ground-truth answer or more strongly associated with the question entity. Moreover, model confidence is closely related to answer-side popularity signals, particularly question–answer co-occurrence, even when the generated answer is incorrect. Together, these findings reveal a popularity-associated pattern of overconfidence, in which popular or strongly associated incorrect alternatives can receive high confidence. We further show that these popularity signals can be leveraged for confidence calibration, substantially mitigating overconfidence while improving overall calibration quality. \looseness=-1

\section*{Limitations}
Our study has several limitations. First, we do not have access to the actual training data of the evaluated LLMs, so our popularity measures serve only as proxies for how frequently particular knowledge may have been encountered during training. Second, our findings are correlational rather than causal: they reveal systematic associations between popularity and overconfidence, but do not establish that popularity itself causes overconfident predictions. Finally, our analysis focuses on closed-book entity-centric factual QA, where entity popularity and question-answer co-occurrence can be measured consistently. Whether these patterns extend to more complex settings, such as multi-hop QA, open-ended generation, and reasoning-intensive tasks, remains an important direction for future work. \looseness=-1

\section*{Ethics Statement}
All datasets and models used are open-source or publicly accessible. This work aims to understand and mitigate LLM overconfidence and does not introduce new harmful capabilities. Wikidata and Wikipedia are used only for entity- and document-level statistics. AI-based tools were used solely for language polishing, grammar correction, and translation; all ideas, experiments, analyses, results, and claims were produced and verified by the authors.

\section*{Acknowledgements}
This work was funded by the National Natural Science Foundation of China (NSFC) under Grants No. 62302486 and No. 62441229, the Innovation Project of ICT CAS under Grants E561010 and No. E361140, and the Strategic Priority Research Program of the CAS under Grants No. XDB0680102.

\bibliography{custom}
\bibliographystyle{acl_natbib}

\appendix

\section{Knowledge Popularity Affects Model Learning \label{sec:acc_learning}}
\paragraph{Co-occurrence frequency exhibits the strongest correlation with model QA performance.} 
On Movies and Songs, $\text{RPop}_\text{GT}$ is the dominant predictor, with partial correlations up to $r = 0.39$, substantially outperforming both question and answer popularity. This result suggests that factual recall is strongly associated with the frequency of (question, answer) co-occurrence in pretraining corpora.
On Basketball, the signal most strongly correlated with accuracy is not $\text{RPop}_\text{GT}$ for either Llama-3-8B or Qwen2-7B. This result may be attributed to their relatively weak performance on this dataset.
As shown in Table~\ref{tab:chatgpt-failure-mode}, unlike the other two datasets, the co-occurrence between basketball players and their places of birth is too sparse, making it difficult for weaker models to learn effectively.

\begin{table}[t]
\centering
\small
\caption{Partial correlations between the three types of popularity and accuracy. We do not include the Qwen2.5 series on Basketball because their accuracy is too low.}
\setlength{\tabcolsep}{4pt}
\begin{tabular}{l ccc}
\toprule
Model & $\text{RPop}_\text{GT}$ & $\text{Pop}_\text{Q}$ & $\text{Pop}_\text{GT}$ \\
\midrule
\multicolumn{4}{l}{\textit{Movies}} \\
Llama-3-8B   & \textbf{+0.216} & +0.196 & $-$0.019 \\
Qwen2-7B     & \textbf{+0.323} & +0.287 & $-$0.039 \\
ChatGPT      & +0.083 & \textbf{+0.094} & $-$0.026 \\
Qwen2.5-7B   & +0.285 & \textbf{+0.315} & $-$0.005 \\
Qwen2.5-14B  & \textbf{+0.323} & +0.270 & $-$0.038 \\
Qwen2.5-32B  & \textbf{+0.347} & +0.319 & $-$0.041 \\
\midrule
\multicolumn{4}{l}{\textit{Songs}} \\
Llama-3-8B   & \textbf{+0.391} & +0.102 & $-$0.001 \\
Qwen2-7B     & \textbf{+0.345} & +0.220 & +0.078 \\
ChatGPT      & \textbf{+0.264} & +0.010 & +0.186 \\
Qwen2.5-7B   & \textbf{+0.213} & +0.196 & $-$0.012 \\
Qwen2.5-14B  & \textbf{+0.322} & +0.219 & +0.058 \\
Qwen2.5-32B  & \textbf{+0.306} & +0.231 & $-$0.028 \\
\midrule
\multicolumn{4}{l}{\textit{Basketball}} \\
Llama-3-8B   & +0.139 & +0.080 & \textbf{+0.259} \\
Qwen2-7B     & +0.073 & +0.000 & \textbf{+0.321} \\
ChatGPT      & \textbf{+0.245} & +0.219 & +0.175 \\
\bottomrule
\end{tabular}
\label{tab:acc-corr}
\end{table}

\begin{figure}
    \centering
    \includegraphics[width=1.0\linewidth]{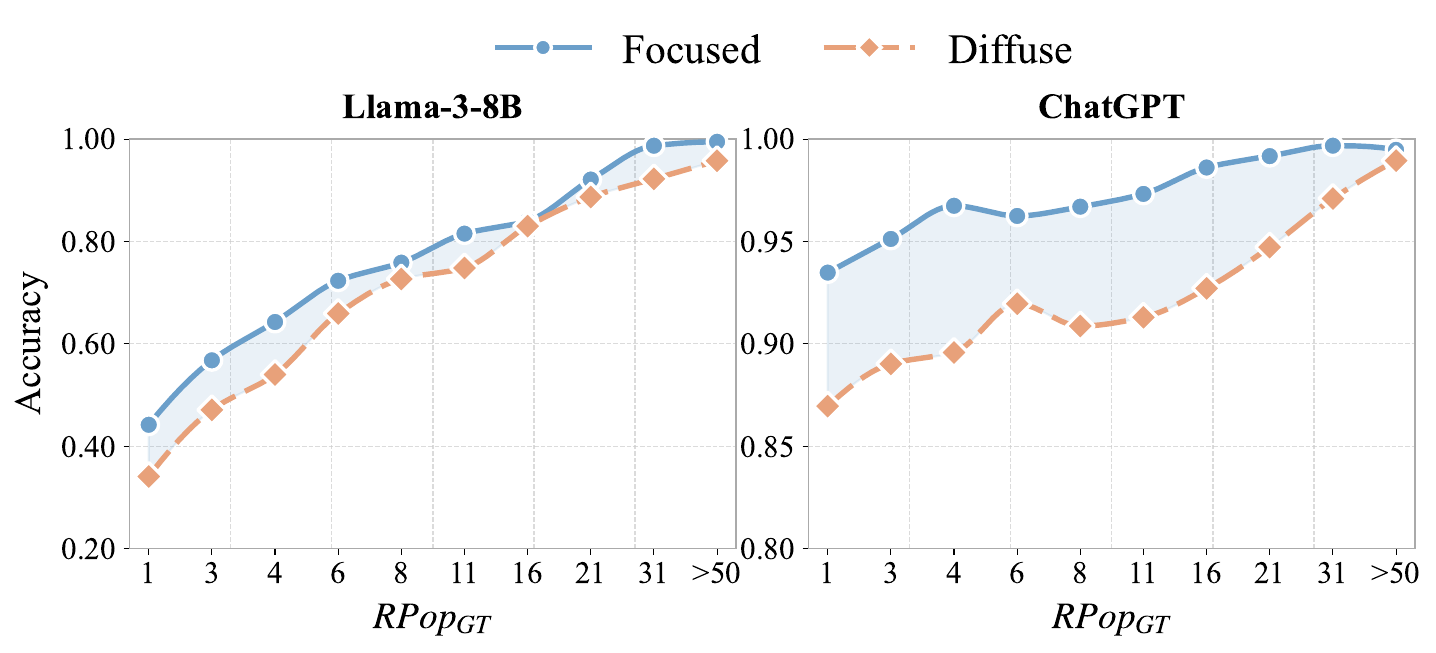}
    \caption{The impact of relation concentration on accuracy. Here, ``focused'' refers to the top half of samples ranked by $\text{RPop}_\text{GT}/\text{Pop}_Q$ within each $\text{RPop}_\text{GT}$ bin, while ``diffuse'' refers to the bottom half of samples.}
    \label{fig:relation_quality}
    \vspace{-0.5cm}
\end{figure}

\paragraph{Relation specificity predicts accuracy beyond co-occurrence.}
We argue that co-occurrence count alone is insufficient to characterize learnability. The same frequency may correspond to either a highly focused relation (e.g., director is the dominant association of a movie) or a diffuse one where the relation is only one of many facts about the entity.
We define relation specificity as $\frac{\text{RPop}_\text{GT}}{\text{Pop}_Q}$, where higher values indicate that the (question, answer) relation is more dominant for the question entity. To control for raw frequency, we bin examples by $\text{RPop}_\text{GT}$ and compare high vs. low specificity within each bin.
As shown in Figure~\ref{fig:relation_quality}, higher specificity consistently yields higher accuracy, suggesting that relation specificity is an important factor in factual learning.

\section{Detailed Results}

\begin{table*}[t]
\centering
\small
\caption{Accuracy and average confidence for each model. We omit results for the Qwen2.5 series on Basketball in subsequent analyses because their extremely low accuracy limits the reliability of the estimates.}
\label{tab:acc_and_conf}
\setlength{\tabcolsep}{4pt}
\begin{tabular}{ll cccccc}
\toprule
Datasets & Metrics & Llama-3-8B & Qwen2-7B & ChatGPT & Qwen2.5-7B & Qwen2.5-14B & Qwen2.5-32B \\
\midrule
\multirow{2}{*}{Movies}
& Acc  & 0.715 & 0.401 & 0.942 & 0.210 & 0.469 & 0.437 \\
& Conf & 0.904 & 0.817 & 0.986 & 0.754 & 0.877 & 0.877 \\
\midrule
\multirow{2}{*}{Songs} 
& Acc  & 0.364 & 0.255 & 0.719 & 0.117 & 0.247 & 0.185 \\
& Conf & 0.790 & 0.777 & 0.947 & 0.788 & 0.865 & 0.844 \\
\midrule
\multirow{2}{*}{Basketball}
& Acc  & 0.134 & 0.098 & 0.350 & 0.042 & 0.074 & 0.078 \\
& Conf & 0.601 & 0.748 & 0.791 & 0.786 & 0.821 & 0.797 \\
\bottomrule
\end{tabular}
\end{table*}

\begin{table*}[t]
\centering
\small
\setlength{\tabcolsep}{3pt}
\caption{Accuracy and popularity-bias ratios across datasets for all models except ChatGPT. Ratios are computed on incorrect predictions. ``Either'' indicates that either $\mathrm{Pop}_{\mathrm{Ge}} > \mathrm{Pop}_{\mathrm{GT}}$ or $\mathrm{RPop}_{\mathrm{Ge}} > \mathrm{RPop}_{\mathrm{GT}}$ holds; ``Pop only'' indicates that only the former holds.}
\label{tab:cross-dataset-all-models}
\begin{tabular}{ll c cccc}
\toprule
& & & \multicolumn{4}{c}{Ratios on incorrect predictions} \\
\cmidrule(lr){4-7}
Dataset & Model & Acc. &
$\mathrm{Pop}_{\mathrm{Ge}} > \mathrm{Pop}_{\mathrm{GT}}$ &
$\mathrm{RPop}_{\mathrm{Ge}} > \mathrm{RPop}_{\mathrm{GT}}$ &
Either & Pop only \\
\midrule
Movies
& Llama-3-8B  & 0.715 & 0.557 & 0.101 & 0.579 & 0.478 \\
& Qwen2-7B    & 0.401 & 0.642 & 0.096 & 0.653 & 0.556 \\
& Qwen2.5-7B  & 0.210 & 0.561 & 0.047 & 0.568 & 0.521 \\
& Qwen2.5-14B & 0.469 & 0.478 & 0.045 & 0.487 & 0.443 \\
& Qwen2.5-32B & 0.437 & 0.551 & 0.057 & 0.564 & 0.508 \\
\midrule
Songs
& Llama-3-8B  & 0.364 & 0.448 & 0.159 & 0.517 & 0.358 \\
& Qwen2-7B    & 0.255 & 0.561 & 0.120 & 0.590 & 0.470 \\
& Qwen2.5-7B  & 0.117 & 0.596 & 0.095 & 0.615 & 0.520 \\
& Qwen2.5-14B & 0.247 & 0.437 & 0.093 & 0.460 & 0.367 \\
& Qwen2.5-32B & 0.185 & 0.516 & 0.091 & 0.535 & 0.444 \\
\midrule
Basketball
& Llama-3-8B  & 0.134 & 0.687 & 0.332 & 0.747 & 0.415 \\
& Qwen2-7B    & 0.098 & 0.782 & 0.424 & 0.843 & 0.419 \\
& Qwen2.5-7B  & 0.042 & 0.731 & 0.473 & 0.829 & 0.356 \\
& Qwen2.5-14B & 0.074 & 0.788 & 0.452 & 0.839 & 0.387 \\
& Qwen2.5-32B & 0.078 & 0.794 & 0.406 & 0.834 & 0.428 \\
\bottomrule
\end{tabular}
\end{table*}

\begin{table*}[t]
\centering
\small
\caption{Generated-answer statistics by correctness. $Pop_{Ge}$ and $RPop_{Ge}$ denote the median entity popularity and question--answer association strength, respectively. CV is reported only for $Pop_{Ge}$ because $RPop_{Ge}$ is highly zero-inflated, making its CV unstable.}
\label{tab:gene-answer-stats}
\setlength{\tabcolsep}{4pt}
\begin{tabular}{ll c cc c c cc c}
\toprule
& & \multicolumn{4}{c}{Correct} & \multicolumn{4}{c}{Incorrect} \\
\cmidrule(lr){3-6}\cmidrule(lr){7-10}
Model & Dataset & N & $Pop_{Ge}$ & $\mathrm{CV}(Pop_{Ge})$ & $RPop_{Ge}$ & N & $Pop_{Ge}$ & $\mathrm{CV}(Pop_{Ge})$ & $RPop_{Ge}$ \\
\midrule
Llama-3-8B & Movies & 6319 & 32.0 & 0.85 & 11.0 & 3205 & 17.0 & 1.22 & 0.0 \\
 & Songs & 417 & 62.0 & 0.69 & 27.0 & 781 & 38.0 & 0.92 & 1.0 \\
 & Basketball & 1771 & 191.0 & 0.47 & 4.0 & 11529 & 149.0 & 0.54 & 1.0 \\
\midrule
Qwen2-7B & Movies & 3678 & 38.0 & 0.77 & 16.0 & 5892 & 31.0 & 0.83 & 1.0 \\
 & Songs & 319 & 80.0 & 0.56 & 38.0 & 1011 & 51.0 & 0.75 & 0.0 \\
 & Basketball & 1305 & 223.0 & 0.35 & 3.0 & 11995 & 191.0 & 0.42 & 2.0 \\
\midrule
ChatGPT & Movies & 8768 & 27.0 & 0.90 & 9.0 & 828 & 5.0 & 1.58 & 2.0 \\
 & Songs & 1034 & 66.0 & 0.63 & 18.0 & 436 & 39.0 & 0.90 & 4.5 \\
 & Basketball & 4633 & 132.0 & 0.66 & 3.0 & 8667 & 186.0 & 0.58 & 1.0 \\
\midrule
Qwen2.5-7B & Movies & 1749 & 45.0 & 0.71 & 24.0 & 7812 & 23.0 & 1.00 & 0.0 \\
 & Songs & 147 & 87.0 & 0.56 & 56.0 & 1205 & 66.0 & 0.71 & 0.0 \\
\midrule
Qwen2.5-14B & Movies & 3881 & 37.0 & 0.78 & 15.0 & 5662 & 15.0 & 1.23 & 0.0 \\
 & Songs & 298 & 81.0 & 0.57 & 43.5 & 1055 & 38.0 & 0.94 & 0.0 \\
\midrule
Qwen2.5-32B & Movies & 3747 & 38.0 & 0.77 & 16.0 & 5771 & 20.0 & 1.09 & 0.0 \\
 & Songs & 241 & 78.0 & 0.58 & 48.0 & 1165 & 50.0 & 0.80 & 0.0 \\
\bottomrule
\end{tabular}

\end{table*}

\begin{table*}[t]
\centering
\small
\setlength{\tabcolsep}{8pt}
\caption{Difference in mean question-generated answer association between correct and incorrect predictions within confidence quartiles for open-source models. Positive values indicate that correct answers have stronger question-answer associations than incorrect answers at similar confidence levels. ``-'' denotes fewer than five incorrect examples; Avg is the mean over available models.}

\label{tab:delta-gcoo}
\begin{tabular}{lcccc}
\toprule
& \multicolumn{4}{c}{
    $\Delta\mathrm{RPop}_{\mathrm{Ge}}$ by confidence quartile
} \\
\cmidrule(lr){2-5}
Model & Q1 (low) & Q2 & Q3 & Q4 (high) \\
\midrule
\multicolumn{5}{l}{\textit{Movies}} \\
\quad Llama-3-8B  & +7.6  & +6.7  & +3.1  & +3.3  \\
\quad Qwen2-7B    & +8.6  & +11.9 & +13.1 & +18.2 \\
\quad Qwen2.5-7B  & +14.8 & +17.5 & +32.6 & +33.1 \\
\quad Qwen2.5-14B & +8.9  & +9.0  & +13.1 & +11.1 \\
\quad Qwen2.5-32B & +8.6  & +10.6 & +15.1 & +14.0 \\
\quad \textbf{Avg}
                  & \textbf{+9.7}
                  & \textbf{+11.1}
                  & \textbf{+15.4}
                  & \textbf{+16.0} \\
\midrule
\multicolumn{5}{l}{\textit{Songs}} \\
\quad Llama-3-8B  & +18.2 & +20.4 & +4.9  & $-$0.1 \\
\quad Qwen2-7B    & +13.0 & +28.2 & +19.2 & +24.6  \\
\quad Qwen2.5-7B  & --    & +48.7 & +54.2 & +24.6  \\
\quad Qwen2.5-14B & --    & +30.0 & +27.4 & +2.8   \\
\quad Qwen2.5-32B & --    & +32.3 & +37.7 & +6.1   \\
\quad \textbf{Avg}
                  & \textbf{+15.6}
                  & \textbf{+31.9}
                  & \textbf{+28.7}
                  & \textbf{+11.6} \\
\midrule
\multicolumn{5}{l}{\textit{Basketball}} \\
\quad Llama-3-8B & +4.0 & +3.7 & +3.8 & +7.0  \\
\quad Qwen2-7B   & +3.6 & +1.3 & +1.0 & $-$1.6 \\
\quad \textbf{Avg}
                 & \textbf{+3.8}
                 & \textbf{+2.5}
                 & \textbf{+2.4}
                 & \textbf{+2.7} \\
\bottomrule
\end{tabular}
\end{table*}

\begin{table*}[t]
\centering
\small
\setlength{\tabcolsep}{8pt}
\caption{Difference in mean generated-answer entity popularity between correct and incorrect predictions within confidence quartiles for open-source models. Positive values indicate that correct answers are more popular than incorrect answers at similar confidence levels. ``-'' denotes fewer than five incorrect examples; Avg is the unweighted mean over available models.}

\label{tab:delta-gpop}
\begin{tabular}{lcccc}
\toprule
& \multicolumn{4}{c}{
    $\Delta\mathrm{Pop}_{\mathrm{Ge}}$ by confidence quartile
} \\
\cmidrule(lr){2-5}
Model & Q1 (low) & Q2 & Q3 & Q4 (high) \\
\midrule
\multicolumn{5}{l}{\textit{Movies}} \\
\quad Llama-3-8B  & $-$6.4  & $-$11.6 & $-$12.8 & +7.2   \\
\quad Qwen2-7B    & $-$10.2 & $-$8.8  & $-$12.3 & $-$5.6 \\
\quad Qwen2.5-7B  & +7.6    & +1.3    & +3.1    & +3.9   \\
\quad Qwen2.5-14B & +7.3    & +3.9    & +2.7    & $-$1.5 \\
\quad Qwen2.5-32B & +1.6    & $-$1.3  & +1.4    & $-$2.5 \\
\quad \textbf{Avg}
                  & \textbf{$-$0.0}
                  & \textbf{$-$3.3}
                  & \textbf{$-$3.6}
                  & \textbf{+0.3} \\
\midrule
\multicolumn{5}{l}{\textit{Songs}} \\
\quad Llama-3-8B  & +30.5 & +22.1  & +4.3   & +39.3  \\
\quad Qwen2-7B    & +17.7 & +5.7   & +13.8  & +8.3   \\
\quad Qwen2.5-7B  & --    & $-$21.5 & $-$13.2 & $-$0.9 \\
\quad Qwen2.5-14B & --    & +14.3  & +11.4  & +13.8  \\
\quad Qwen2.5-32B & --    & $-$9.4 & +1.1   & +2.5   \\
\quad \textbf{Avg}
                  & \textbf{+24.1}
                  & \textbf{+2.2}
                  & \textbf{+3.5}
                  & \textbf{+12.6} \\
\midrule
\multicolumn{5}{l}{\textit{Basketball}} \\
\quad Llama-3-8B & +38.0 & +48.0 & +38.7 & +11.0 \\
\quad Qwen2-7B   & +17.7 & +38.0 & +25.7 & +17.4 \\
\quad \textbf{Avg}
                 & \textbf{+27.9}
                 & \textbf{+43.0}
                 & \textbf{+32.2}
                 & \textbf{+14.2} \\
\bottomrule
\end{tabular}
\end{table*}

\subsection{Per-Model Detection Results}
\label{sec:per-model-detect}

Tables~\ref{tab:detect-per-model-movies}--\ref{tab:detect-per-model-basketball} report per-model Alignment on Movies, Songs, and Basketball, respectively. The pattern described in \S\ref{sec:rq3-detect} holds consistently across all six models: $\text{RPop}_\text{Ge}$ alone outperforms all confidence baselines on Movies and Songs, and Prob~+~$\text{RPop}_\text{Ge}$ (MLP) yields the best single-feature result on Movies and Songs, while $\text{Pop}_\text{Ge}$ provides an effective complementary signal on Basketball.

\begin{table*}[t]
\centering
\small
\setlength{\tabcolsep}{4pt}
\caption{Per-model Alignment (\%) on Movies. Boldface marks the best method within each model column. All ext.\ = $\{\text{Pop}_\text{Q}, \text{Pop}_\text{Ge}, \text{RPop}_\text{Ge}\}$; All LLM = $\{\widehat{\text{Pop}}_\text{Q}, \widehat{\text{Pop}}_\text{Ge}, \widehat{\text{RPop}}_\text{Ge}\}$.}
\label{tab:detect-per-model-movies}
\begin{tabular}{l cccccc}
\toprule
Method & Llama-3-8B & Qwen2-7B & ChatGPT & Qwen2.5-7B & Qwen2.5-14B & Qwen2.5-32B \\
\midrule
\multicolumn{7}{l}{\textit{Confidence alone}} \\
\quad Prob & 83.2 & 81.1 & 78.6 & 82.0 & 81.7 & 84.7 \\
\quad Consis & 74.1 & 81.4 & 70.9 & 82.1 & 83.7 & 83.7 \\
\quad Verbal & 75.5 & 54.7 & 58.2 & 65.5 & 79.4 & 58.3 \\
\quad Prob + Temp & 83.1 & 81.1 & 78.6 & 82.0 & 81.7 & 84.7 \\
\midrule
\multicolumn{7}{l}{\textit{Corpora-based popularity}} \\
\quad $\text{Pop}_\text{Q}$ & 66.9 & 72.6 & 62.6 & 74.3 & 72.1 & 73.1 \\
\quad $\text{Pop}_\text{Ge}$ & 74.4 & 62.5 & 71.7 & 65.5 & 67.3 & 64.6 \\
\quad $\text{RPop}_\text{Ge}$ & 89.3 & 87.6 & 71.6 & 93.5 & 92.2 & 91.7 \\
\midrule
\multicolumn{7}{l}{\textit{Popularity + confidence (MLP)}} \\
\quad Prob + $\text{Pop}_\text{Q}$ & 83.9 & 83.2 & 78.6 & 85.0 & 83.2 & 85.4 \\
\quad Prob + $\text{Pop}_\text{Ge}$ & 85.1 & 81.5 & 81.0 & 81.9 & 81.9 & 84.7 \\
\quad Prob + $\text{RPop}_\text{Ge}$ & 90.3 & 88.0 & 80.2 & 93.6 & 92.4 & 92.4 \\
\quad Prob + All ext.\ & 92.9 & 92.0 & 82.5 & 94.7 & 93.2 & 93.7 \\
\midrule
\multicolumn{7}{l}{\textit{Self-generated popularity (MLP)}} \\
\quad Prob + $\widehat{\text{RPop}}_\text{Ge}$ & 85.3 & 81.2 & 78.7 & 84.0 & 84.2 & 85.1 \\
\quad Prob + All LLM & 85.9 & 81.1 & 75.2 & 86.8 & 85.4 & 87.1 \\
\bottomrule
\end{tabular}
\end{table*}

\begin{table*}[t]
\centering
\small
\setlength{\tabcolsep}{4pt}
\caption{Per-model Alignment (\%) on Songs. Boldface marks the best method within each model column.}
\label{tab:detect-per-model-songs}
\begin{tabular}{l cccccc}
\toprule
Method & Llama-3-8B & Qwen2-7B & ChatGPT & Qwen2.5-7B & Qwen2.5-14B & Qwen2.5-32B \\
\midrule
\multicolumn{7}{l}{\textit{Confidence alone}} \\
\quad Prob & 77.9 & 78.7 & 76.0 & 85.1 & 84.7 & 81.8 \\
\quad Consis & 79.1 & 85.8 & 80.5 & 84.5 & 89.0 & 82.2 \\
\quad Verbal & 62.5 & 54.3 & 74.9 & 65.5 & 79.4 & 57.9 \\
\quad Prob + Temp & 77.9 & 78.7 & 76.2 & 85.1 & 84.7 & 81.8 \\
\midrule
\multicolumn{7}{l}{\textit{Corpora-based popularity}} \\
\quad $\text{Pop}_\text{Q}$ & 69.8 & 68.8 & 68.0 & 64.2 & 71.8 & 67.4 \\
\quad $\text{Pop}_\text{Ge}$ & 67.9 & 62.0 & 71.0 & 54.7 & 67.8 & 55.8 \\
\quad $\text{RPop}_\text{Ge}$ & 81.6 & 88.6 & 78.3 & 91.2 & 93.4 & 90.5 \\
\midrule
\multicolumn{7}{l}{\textit{Popularity + confidence (MLP)}} \\
\quad Prob + $\text{Pop}_\text{Q}$ & 79.8 & 82.1 & 77.1 & 83.8 & 88.7 & 82.2 \\
\quad Prob + $\text{Pop}_\text{Ge}$ & 79.4 & 79.6 & 76.6 & 86.5 & 85.0 & 83.1 \\
\quad Prob + $\text{RPop}_\text{Ge}$ & 84.4 & 89.5 & 79.3 & 93.9 & 94.0 & 91.3 \\
\quad Prob + All ext.\ & 83.3 & 87.7 & 79.7 & 94.6 & 93.4 & 88.8 \\
\midrule
\multicolumn{7}{l}{\textit{Self-generated popularity (MLP)}} \\
\quad Prob + $\widehat{\text{RPop}}_\text{Ge}$ & 80.7 & 79.9 & 76.0 & 87.8 & 88.0 & 83.1 \\
\quad Prob + All LLM & 81.2 & 80.2 & 77.3 & 86.5 & 88.0 & 87.6 \\
\bottomrule
\end{tabular}
\end{table*}

\begin{table*}[t]
\centering
\small
\setlength{\tabcolsep}{4pt}
\caption{Per-model Alignment (\%) on Basketball. Boldface marks the best method within each model column.}
\label{tab:detect-per-model-basketball}
\begin{tabular}{l cccccc}
\toprule
Method & Llama-3-8B & Qwen2-7B & ChatGPT & Qwen2.5-7B & Qwen2.5-14B & Qwen2.5-32B \\
\midrule
\multicolumn{7}{l}{\textit{Confidence alone}} \\
\quad Prob & 66.3 & 68.3 & 77.7 & 62.9 & 66.9 & 69.8 \\
\quad Consis & 66.8 & 61.0 & 79.1 & 63.4 & 62.4 & 70.2 \\
\quad Verbal & 62.7 & 51.4 & 55.8 & 65.5 & 50.0 & 62.2 \\
\quad Prob + Temp & 66.3 & 68.3 & 77.7 & 63.2 & 66.9 & 70.1 \\
\midrule
\multicolumn{7}{l}{\textit{Corpora-based popularity}} \\
\quad $\text{Pop}_\text{Q}$ & 57.1 & 51.5 & 69.6 & 48.3 & 52.0 & 51.6 \\
\quad $\text{Pop}_\text{Ge}$ & 60.0 & 61.1 & 65.2 & 61.6 & 57.5 & 63.1 \\
\quad $\text{RPop}_\text{Ge}$ & 66.3 & 61.0 & 65.2 & 56.5 & 59.4 & 63.9 \\
\midrule
\multicolumn{7}{l}{\textit{Popularity + confidence (MLP)}} \\
\quad Prob + $\text{Pop}_\text{Q}$ & 67.1 & 67.9 & 78.1 & 62.5 & 65.7 & 68.7 \\
\quad Prob + $\text{Pop}_\text{Ge}$ & 68.5 & 69.0 & 78.3 & 72.5 & 68.2 & 73.4 \\
\quad Prob + $\text{RPop}_\text{Ge}$ & 71.5 & 69.0 & 78.1 & 63.4 & 66.5 & 71.3 \\
\quad Prob + All ext.\ & 70.1 & 69.0 & 78.3 & 70.9 & 69.2 & 73.9 \\
\midrule
\multicolumn{7}{l}{\textit{Self-generated popularity (MLP)}} \\
\quad Prob + $\widehat{\text{RPop}}_\text{Ge}$ & 66.3 & 69.0 & 77.5 & 61.6 & 66.1 & 70.0 \\
\quad Prob + All LLM & 66.4 & 69.2 & 77.8 & 68.2 & 68.0 & 72.1 \\
\bottomrule
\end{tabular}
\end{table*}

\subsection{Training and Evaluation Details}
\label{sec:training-details}

\paragraph{Data preparation.} For each (dataset, model) pair, we collect samples from the model's temperature-1.0 generations. We filter out samples where the model produces no answer. For Movies and Songs, we additionally discard samples where the question or generated entity has a Wikidata single-occurrence count above 6{,}000, as such ambiguous high-frequency entity names introduce noise. External popularity and co-occurrence values that cannot be found in Wikidata/Wikipedia are set to zero rather than discarded, reflecting the deployment scenario where not all entities appear in the knowledge base.

\paragraph{Class balancing.} When the accuracy of a (dataset, model) pair falls outside $[0.25, 0.75]$, we undersample the majority class to achieve a 1:1 positive-to-negative ratio \emph{before} the train/test split. This ensures that no method---including a trivial majority-class predictor---can achieve high Alignment from class imbalance alone. For models whose QA accuracy already falls within this range (e.g., Qwen2-7B on Movies, accuracy $\approx 0.38$), no undersampling is applied.

\paragraph{Evaluation settings.} The main setting applies the conditional class-balancing procedure described above before the train/test split. The natural-distribution setting retains the original class distribution without undersampling. In both settings, the calibration methods and MLPs are fitted on the corresponding training split and evaluated on the held-out test split.

\paragraph{Train/test split.} The (possibly balanced) samples are randomly split 50/50 into train and test sets with stratified sampling (seed 42). All methods use the same split to ensure fair comparison.

\paragraph{Calibration fitting and threshold selection.} All fitting and threshold selection are performed independently for each (dataset, model) pair using only the training split; the held-out test split is used solely for evaluation. For threshold-based methods (Prob, Consis, Verbal, and standalone popularity signals), the optimal binarization threshold is selected by scanning all unique training-set values and choosing the one that maximizes training-set accuracy. For temperature scaling, Platt scaling, and isotonic regression, the calibration mapping is first fitted on the training split, after which the optimal threshold is selected on the corresponding calibrated training-set probabilities.

\paragraph{MLP training.} The MLP uses three hidden layers (128, 64, 32 units), each followed by batch normalization, ReLU activation, and dropout (rate 0.3). It is trained with Adam (learning rate $5 \times 10^{-4}$, weight decay $10^{-4}$) and binary cross-entropy loss. For each (dataset, model) pair, 10\% of its training split is held out as a validation subset (stratified, seed 42). We retain the checkpoint with the lowest validation BCE loss and stop training after 10 consecutive epochs without improvement. After training, the optimal binarization threshold is selected on the full training split, and Alignment is evaluated on the held-out test split. In transfer experiments, checkpoint selection uses only the validation subset of the source pair; no target-test samples are used for training or model selection.

\subsection{Full Calibration Transfer Results}
\label{sec:full-transfer-results}

For cross-dataset transfer, each method is fitted on the training split of a source dataset and evaluated on the test split of a different target dataset for the same model. Results are macro-averaged over 36 ordered source--target pairs. For cross-model transfer, each method is fitted on a source model and evaluated on a different target model within the same dataset, with results macro-averaged over 90 ordered pairs. All fitted parameters, feature scalers, and decision thresholds come exclusively from the source training split; target labels are used only for evaluation. The main setting applies the conditional balancing procedure in Appendix~\ref{sec:training-details}, whereas the natural-distribution setting retains the original class distribution for both source and target pairs. \looseness=-1

\subsection{Partial Correlation Computation}
\label{sec:partial-corr-details}

All partial correlation coefficients in this paper are computed using the rank-space OLS residual method, which yields the partial Spearman correlation:

\begin{enumerate}[leftmargin=*,topsep=2pt,itemsep=1pt]
\item \textbf{Rank transformation.} Convert each variable (predictor $x$, target $y$, and control variables $\mathbf{z} = \{z_1, \ldots, z_k\}$) to ranks.
\item \textbf{Residualize.} Regress each ranked variable on the ranked controls via ordinary least squares:
\begin{align}
\hat{\beta}_x &= \arg\min_{\beta} \| \mathrm{rank}(x) - \mathbf{Z}\beta \|_2^2, \\
\hat{\beta}_y &= \arg\min_{\beta} \| \mathrm{rank}(y) - \mathbf{Z}\beta \|_2^2,
\end{align}
where $\mathbf{Z} = [\mathrm{rank}(z_1), \ldots, \mathrm{rank}(z_k)]$ is the matrix of ranked control variables (mean-centered). The residuals
\begin{align}
r_x &= \mathrm{rank}(x) - \mathbf{Z}\hat{\beta}_x, \\
r_y &= \mathrm{rank}(y) - \mathbf{Z}\hat{\beta}_y
\end{align}
represent the parts of $x$ and $y$ not linearly explained by the controls, in the rank space.
\item \textbf{Correlate.} The partial Spearman correlation is $r = \mathrm{Pearson}(r_x, r_y)$.
\end{enumerate}

The three-variable mutual-control framework used throughout the paper selects three factors $\{f_1, f_2, f_3\}$ and, for each $f_i$, computes the partial correlation between $f_i$ and the target while controlling for $\{f_j : j \neq i\}$. This ensures that each reported coefficient captures the unique contribution of that factor after removing the linear influence of the other two. Two variable sets are used depending on the type of knowledge:
\begin{itemize}[leftmargin=*,topsep=2pt,itemsep=1pt]
\item Ground-truth-side knowledge: $\{\text{RPop}_\text{GT}, \text{Pop}_\text{Q}, \text{Pop}_\text{GT}\}$, where $\text{RPop}_\text{GT}$ is the Wikipedia co-occurrence count of the question entity and ground-truth answer, $\text{Pop}_\text{Q}$ is the Wikidata sitelink count of the question entity, and $\text{Pop}_\text{GT}$ is the Wikidata sitelink count of the ground-truth answer.
\item Generation-side knowledge: $\{\text{RPop}_\text{Ge}, \text{Pop}_\text{Q}, \text{Pop}_\text{Ge}\}$, where $\text{RPop}_\text{Ge}$ and $\text{Pop}_\text{Ge}$ replace their ground-truth counterparts with the generated answer's equivalents, since the model can only observe its own output.
\end{itemize}
All analyses are performed independently per (dataset, model) pair.

\end{document}